\documentclass{article}
\usepackage{graphicx} 
\usepackage{amsmath,amssymb}
\usepackage{lipsum}
\usepackage{newpxtext}
\usepackage{tcolorbox}
\usepackage{xcolor}
\usepackage{color}
\usepackage{booktabs}
\usepackage{parskip}
\usepackage[numbers]{natbib}
\usepackage{sectsty}
\usepackage{authblk}
\usepackage{subcaption}
\usepackage[left=1.5in, right=1.5in, top=1.5in, bottom=1.5in]{geometry}
\usepackage{hyperref}
\hypersetup{colorlinks = true, linkcolor = black,
            urlcolor  = gray,
            citecolor = violet,
            anchorcolor = brown}
\newtcolorbox{blockquote}[1][]{
    colback=blue!4!white,  
    colframe=blue!75!black,  
    fonttitle=\bfseries,
}
\sectionfont{\fontsize{14pt}{14pt}\selectfont}
\newtcolorbox{blockcomment}[1][]{
    colback=green!4!white,  
    colframe=green!75!black,  
    fonttitle=\bfseries,
}
\newtcolorbox{takeaway}[1][]{
    colback=black!3!white,
    colframe=black!35!white,
    boxrule=0.4pt,
    arc=1mm,
    left=1.2mm,
    right=1.2mm,
    top=1mm,
    bottom=1mm,
}
\newtcolorbox{positionbox}[1][]{
    colback=blue!3!white,
    colframe=blue!45!black,
    boxrule=0.5pt,
    arc=1mm,
    left=1.5mm,
    right=1.5mm,
    top=1.2mm,
    bottom=1.2mm,
}

\usepackage{amsmath,amsthm,amsfonts,bm,physics,mathtools}
\usepackage{cancel}
\usepackage{derivative}

\DeclareMathAlphabet{\mathmybb}{U}{bbold}{m}{n}

\def\rvx{{\mathbf{x}}}

\DeclareMathAlphabet{\mathsfit}{\encodingdefault}{\sfdefault}{m}{sl}
\SetMathAlphabet{\mathsfit}{bold}{\encodingdefault}{\sfdefault}{bx}{n}

\title{Ideas in Inference-time Scaling can Benefit Generative Pre-training Algorithms}
\author{Jiaming Song, Linqi Zhou\thanks{Part of this work was done while the authors were employed at Luma AI; the draft was completed with Codex.}}
\date{}

\begin{document}

\maketitle
\begin{abstract}
    Generative pre-training is often framed through a false dichotomy between autoregressive models for discrete signals and diffusion models for continuous signals. We argue that the dichotomy is false because it conflates model family, data representation, training objective, and inference procedure. Autoregression is an inference procedure that expands a sequence through normalized conditional draws, while diffusion is a refinement procedure that repeatedly revises an existing state. The more useful contrast is therefore not autoregressive versus diffusion, but discrete tokens learned with cross-entropy versus continuous tokens learned with diffusion-style objectives, together with the inference algorithms used to sample from them. From this perspective, algorithmic progress should prioritize inference-time efficiency along two axes: sequence expansion and state refinement. We advocate designing the inference procedure before the training objective, because a training method cannot compensate for an inference map that omits necessary arguments or imposes an incorrect factorization. We illustrate this principle through a target-time limitation of DDIM-style samplers, a joint-distribution limitation of multi-token prediction, and recent flow-map and few-step distillation methods that directly parameterize long-range inference moves.
\end{abstract}
\section{Introduction}

\paragraph{The false dichotomy.}
Despite the significant advances in applications from generative pre-training of foundation models, algorithmic developments in this area have stagnated. The field is often described as being dominated by two paradigms, both popularized around 2020: autoregressive models for discrete signals~\citep{brown2020language} and diffusion models for continuous signals~\citep{ho2020denoising}. This description is useful historically, but it is conceptually misleading. Autoregressive generation and diffusion-style generation are both inference-time scalable algorithms. The former usually spends more compute by generating more tokens; the latter usually spends more compute by refining existing tokens or states. Therefore, the sharper dichotomy is not \textit{autoregressive versus diffusion}, but \textit{discrete tokens learned with cross-entropy versus continuous tokens learned with diffusion}.

\paragraph{Inference-time axes.}
This position paper argues that ideas in inference-time scaling can benefit generative pre-training algorithms. We consider two axes of inference-time scaling: \textit{sequence expansion} and \textit{state refinement}. These axes are commonly associated with autoregressive models and diffusion models respectively, but they are not exclusive to either family. One can combine sequence expansion and state refinement, as in blockwise or hybrid methods, or ask whether a method that currently scales along one axis could be made efficient along the other. Based on this perspective, we believe that practical generative pre-training algorithms should (1) scale \textit{efficiently} along the relevant inference axes, and (2) design the inference algorithm before the training algorithm so that model capacity is used where it matters at inference time.

\paragraph{A continuous-space example.}
We use Inductive Moment Matching (IMM,~\citep{zhou2025inductive}), developed around the same time as this position, as an early concrete example supporting these beliefs in the continuous space. In particular, we examine the one-step iterative process of DDIM~\citep{song2020denoising,lipman2022flow,liu2022flow} and show that it has limited capacity with respect to the target timestep under the current denoising network design. This can be addressed by adding the target timestep to the inputs of the denoising network~\citep{kim2023consistency}.

Interestingly, this one fix, plus a proper moment matching objective~\citep{gretton2012kernel}, leads to a stable, single-stage algorithm that surpasses diffusion models in sample quality while being over an order of magnitude more efficient at inference~\citep{zhou2025inductive}. Notably, these ideas do not rely on denoising score matching~\citep{vincent2011connection} or the score-based stochastic differential equations~\citep{song2020score} on which the foundations of diffusion models are built.

\paragraph{Evidence in hindsight.}
Subsequent progress suggests that this was not an isolated observation. A particularly important line is the flow-map program initiated by Boffi, Albergo, and Vanden-Eijnden, which made explicit that a few-step generator should learn a two-time transport map rather than only an infinitesimal velocity field~\citep{boffi2024flowmap}. Flow maps, shortcut models, mean flows, trajectory distribution matching, and continuous or categorical diffusion language models all move toward parameterizing longer-range inference moves directly~\citep{boffi2024flowmap,frans2024shortcut,geng2025meanflows,luo2025tdm,roos2026categorical,chen2026langflow}. Rather than treating a diffusion or flow model as only a local vector field that must be numerically integrated, recent work directly learns maps between arbitrary noise levels or trajectory endpoints~\citep{boffi2024flowmap,frans2024shortcut,boffi2025selfdistillation,sabour2025align,geng2025meanflows}. These methods differ substantially in training objective, teacher usage, and target domain, but they share the same inference-first principle emphasized here: parameterize the sampler so that long jumps belong to the model class, rather than expecting local denoising to serve as few-step generation.

\begin{takeaway}
\textbf{Reader's map.} Section~2 defines the two inference axes. Section~3 separates autoregression from next-token prediction and diffusion from state refinement. Section~4 applies the lens to DDIM, multi-token prediction, and flow-map methods.
\end{takeaway}



\section{Two axes of inference-time scaling}
\begin{description}
    \item[Sequence expansion] increases the number of generated tokens, positions, blocks, or reasoning steps. This is seen in standard autoregressive large language models (LLMs), where longer chain-of-thought traces, larger generation budgets, or repeated samples can spend more test-time compute~\citep{wei2022chain,snell2024scaling,brown2024large}.
    \item[State refinement] repeatedly updates an existing object without necessarily increasing its length. This is seen in score-based diffusion models, where sampling, denoising, reverse-process, solver, or inference steps reduce discretization error and move a noisy state toward the data distribution~\citep{ho2020denoising,song2020denoising,salimans2022progressive,ma2025inference}.
\end{description}

Terminology varies across communities. The language-modeling literature often describes the first axis through test-time compute, token budgets, generation length, reasoning trace length, long chain-of-thought, or repeated sampling~\citep{wei2022chain,snell2024scaling,brown2024large}. The diffusion literature usually describes the second axis through sampling steps, denoising steps, reverse steps, solver steps, or inference steps~\citep{ho2020denoising,song2020denoising,ma2025inference}. We use \textit{sequence expansion} and \textit{state refinement} as umbrella terms because they describe the inference operation rather than a particular model family.

\begin{takeaway}
\textbf{Takeaway.} Sequence expansion grows the object being sampled; state refinement revises the current object. Autoregressive and diffusion models are familiar examples, not the only possible occupants of these axes.
\end{takeaway}




Under this definition, autoregressive models usually scale through sequence expansion and diffusion models usually scale through state refinement. We list a few notable examples that cover most of the techniques surrounding discrete and continuous signals.

Methods that are not scalable in either sequence expansion or state refinement:
\begin{itemize}
    \item VAE~\citep{kingma2013auto}, GAN~\citep{goodfellow2014generative}, Normalizing Flows~\citep{rezende2015variational}.
\end{itemize}

Methods that are scalable in sequence expansion but not state refinement:
\begin{itemize}
    \item GPT~\citep{brown2020language}, PixelCNN~\citep{oord2016pixel}, MaskGiT~\citep{chang2022maskgit}, VAR~\citep{tian2025visual}.
\end{itemize}

Methods that are scalable in state refinement but not in sequence expansion:
\begin{itemize}
    \item Diffusion models~\citep{ho2020denoising}, Energy-based models~\cite{du2019implicit}, Consistency models~\citep{song2023consistency}.
    \item Parallel non-linear equation solving for autoregressive models~\citep{song2021accelerating}. This uses iterative updates to sample all tokens in parallel, despite being trained with autoregressive objectives.
\end{itemize}

Methods that are scalable in both, with sequence expansion in the outer loop:
\begin{itemize}
    \item AR-Diffusion~\citep{wu2023ar}, Rolling diffusion~\citep{ruhe2024rolling}, MAR~\citep{li2025autoregressive}.
    \item Blockwise parallel decoding~\citep{stern2018blockwise}, which applies ``predict, verify, accept'' as part of the refinement process.
\end{itemize}

Methods that are scalable in both, with state refinement in the outer loop:
\begin{itemize}
    \item Autoregressive distribution smoothing~\citep{meng2021improved}, which performs an iterative denoising process with an autoregressive model as the inner loop.
\end{itemize}

\section{The false dichotomy}

The taxonomy above is useful, but it should not be read as an ``autoregressive versus diffusion'' dichotomy. Autoregression is primarily a factorization and inference procedure: sample a conditional distribution, append a variable, and repeat. Next-token prediction is the usual training objective that makes this procedure practical in language, but the same autoregressive inference principle can be applied to visual, continuous, or partially denoised representations~\citep{gao2025dar}. Diffusion, by contrast, specifies a refinement process over an existing state. It is often implemented non-autoregressively in continuous domains, but that is a design choice rather than a categorical necessity.

This distinction matters because autoregression and diffusion answer different questions. Autoregression asks how to decompose a joint distribution into a sequence of conditional draws; diffusion asks how to transform a simple or corrupted state into a data-like state through iterative refinement. Therefore, the more meaningful comparison is not ``autoregressive or diffusion'', but \textit{discrete tokens learned with cross-entropy} versus \textit{continuous tokens learned with diffusion}. In the familiar discrete-token case, next-token generation has a useful stability property: using more inference compute usually means continuing the sequence, and each continuation step is still a well-defined conditional prediction given the prefix so far. Diffusion-style refinement uses more inference compute differently: it repeatedly changes the current state. This can support parallel correction, but the intermediate states must stay on a path that leads to high-quality data; hence the importance of inference parameterization, step size, and flow-map capacity.

Recent hybrid methods make this false dichotomy increasingly explicit. Diffusion via Autoregressive Models recasts visual diffusion into a next-token prediction problem over tokens aligned with denoising progress~\citep{gao2025dar}. Block Diffusion and Diffusion Forcing combine autoregressive structure with within-block or full-sequence denoising~\citep{arriola2025block,chen2024diffusionforcing}. These examples suggest that autoregression is a stable inference-time scaling mechanism and diffusion is a refinement mechanism. They are not mutually exclusive tribes. Once the concepts are separated, the goal becomes clearer: design pre-training algorithms whose inference procedures scale efficiently through sequence expansion, state refinement, or both.

\begin{takeaway}
\textbf{Takeaway.} The real contrast is discrete tokens learned with cross-entropy versus continuous tokens learned with diffusion. Autoregression is an inference procedure for many representations; diffusion is a refinement procedure that can be paired with or without autoregressive structure.
\end{takeaway}



\section{Designing algorithms that can scale efficiently}

\begin{positionbox}
\textbf{Framing position.} The central contrast is not autoregressive versus diffusion as mutually exclusive model families, but discrete tokens learned with cross-entropy versus continuous tokens learned with diffusion.

\textbf{Design positions.}
\begin{enumerate}
    \item Pre-training algorithms for generative AI should have inference-time scalability through sequence expansion and state refinement.
    \item These algorithms should also scale efficiently with practical step counts.
    \item Before developing the training method, it should be verified whether the model has enough capacity to represent the target distribution during inference.
\end{enumerate}
\end{positionbox}

\paragraph{Why this matters.}
In the visual generative domain, recent work has continued to revisit traditional ideas such as GANs~\citep{kang2023scaling,zhu2023exploring,huang2025gan} and Normalizing Flows~\citep{kolesnikov2024jet,zhai2024normalizing}, despite the overwhelming popularity of diffusion models. We do not dismiss or promote a training design merely because it is adversarial, flow-based, or denoising-based. Instead, we ask what its inference procedure can scale.

This inference view also clarifies the language-modeling case. Scaling through sequence expansion is familiar to LLM practitioners, but scaling through state refinement~\citep{lou2023discrete,shi2024simplified,sahoo2025simple,gat2024discrete} is less settled. This does not mean state refinement is the wrong path for text. It means the inference algorithm must be expressive enough to use the capacity provided by training.

\paragraph{Efficient scaling.}
The second design position concerns latency: if a model can scale at inference time, can it do so with fewer steps? Diffusion distillation reduces the number of continuous-domain refinement updates~\citep{salimans2022progressive}. Diffusion language models in discrete space can potentially reduce the number of decoding steps needed to represent the same output length~\citep{lou2023discrete,shi2024simplified,ou2024your,sahoo2025simple}. Both directions are valuable because they try to preserve inference-time scaling while reducing wall-clock cost.

\paragraph{Inference capacity.}
The third design position is a necessary condition for training to be meaningful. Before choosing an objective, we should check whether the inference process can represent the target family of distributions in the relevant state space. If the universal approximation theorem~\citep{hornik1989multilayer} applies to the network but the sampler itself omits a necessary argument or imposes a factorization assumption, then no training algorithm can solve the problem perfectly. Unless we accept those limitations, we should redesign the algorithm before training.

\begin{takeaway}
\textbf{Section roadmap.} We use three cases to expose the inference object before the training objective. DDIM asks whether a few-step continuous sampler has the right arguments. Multi-token prediction asks whether parallel token prediction represents a joint distribution. Flow maps show how later work makes long-range updates learnable.
\end{takeaway}

\begin{figure}
    \centering
    \begin{subfigure}[t]{0.44\textwidth}
    \vspace{0pt}
        \centering
        \includegraphics[width=\textwidth]{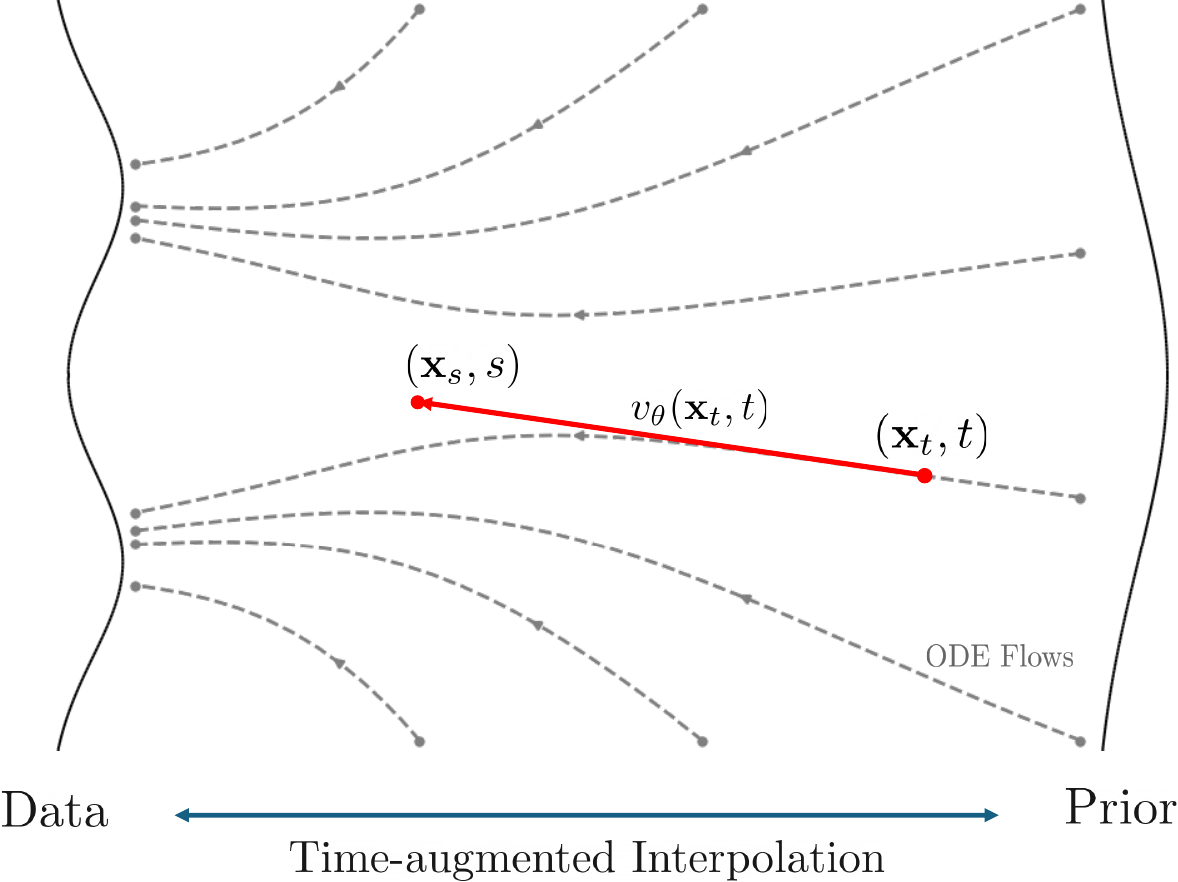}
        \caption{DDIM}
    \end{subfigure} \hspace{5mm}
    \begin{subfigure}[t]{0.44\textwidth}
    \vspace{0pt}
        \centering
        \includegraphics[width=\textwidth]{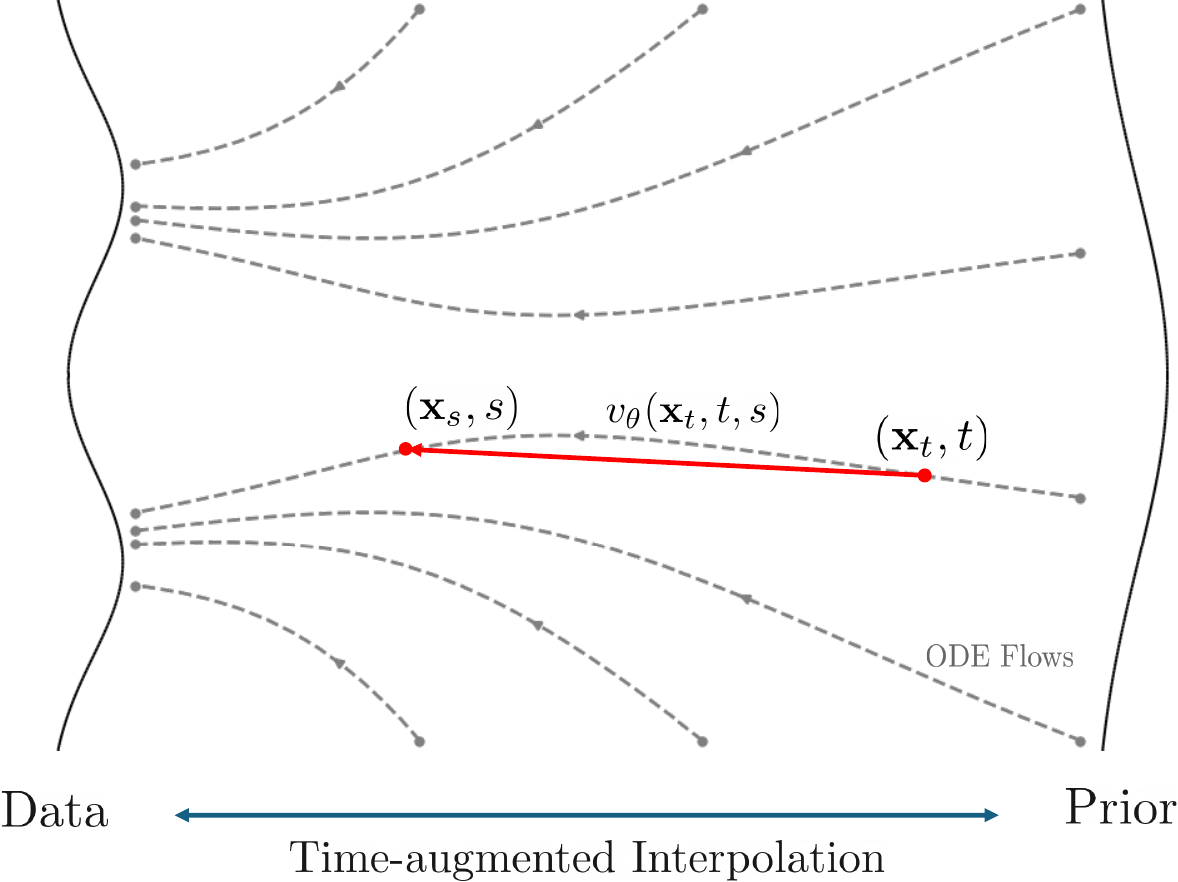}
        \caption{Improved DDIM}
    \end{subfigure}
    \caption{Left shows traditional one-step DDIM sampling under Flow Matching construction. Given $\rvx_t$ and $t$, $\rvx_s$ is produced by following the predicted velocity field $v_\theta(\rvx_t,t)$. However, the model does not have enough capacity to land on the ODE flow result at $s$ in one step because $v_\theta(\rvx_t, t)$ is unaware of $s$ and thus cannot approximate any function over $s$. A practical fix, on the right, simply injects $s$ into our network and now the model has enough capacity to approximate a direct jump towards the correct solution.}
    \label{fig:ddim}
\end{figure}

\subsection{A limitation of DDIM and its improvement}
\begin{takeaway}
\textbf{Core issue.} A DDIM-style one-step update asks the model to jump to a target time $s$, but the usual velocity network does not receive $s$ as an input.
\end{takeaway}

We use the DDIM sampler~\citep{song2020denoising} under rectified flow formulation~\citep{liu2022flow,gao2025diffusionmeetsflow} as an example\footnote{A similar argument would work for any other first-order Euler ODE solver for diffusion models as well.}. In rectified flows, the sampling procedure in each iteration from the current sample $\rvx_t$ and source timestep $t$ to the target timestep $s$ is defined as:
\begin{align}
    \rvx_s := \mathrm{DDIM}(\rvx_t, t, s) = \rvx_t + (s - t) v_\theta(\rvx_t, t),
\end{align}
which transforms the sample $\rvx_t$ to $\rvx_s$ via the velocity network $v_\theta$. Note that in the typical diffusion model formulation, $v_\theta$ takes in only $\rvx_t$ and $t$ as arguments.

\paragraph{A capacity test.}
The denoising autoencoder objective~\citep{vincent2011connection} will not recover a desirable one-step or few-step sampler from this DDIM parameterization. The same issue also constrains other objectives, including diffusion distillation objectives, because the limitation is in the inference map itself.

We believe that the answer is no. To see this, let us take the following partial derivative:
\begin{align}
    \frac{\partial \mathrm{DDIM}(\rvx_t, t, s)}{\partial s} = v_\theta(\rvx_t, t),
\end{align}
which does not depend on $s$ at all. This means that even though DDIM has enough capacity to represent any function over  $\rvx_t$ and $t$ (via universal function approximation), the same does not apply to $s$. We show an illustration of this problem in Figure~\ref{fig:ddim}.

\paragraph{Why training alone cannot fix it.}
This criticism applies to the inference process, not only to how the model is trained. Even if the model is learned via consistency training~\citep{song2023consistency} and can generate accurate samples in a single step, it will not necessarily generate accurate samples under multi-step DDIM. This partially explains why consistency models use restart sampling~\citep{xu2023restart}, which injects additional Gaussian noise at every sampling step.

\paragraph{The minimal fix.}
A natural fix to the above problem is simply to add $s$ also to the input of the velocity network~\citep{kim2023consistency}, so that $v_\theta$ takes three arguments: $\rvx_t$, $t$, and $s$. Now the improved DDIM has enough model capacity to represent functions over $s$ and can learn the proper one-step jump to a solution.

Recently, Zhou \textit{et al.}~\citep{zhou2025inductive} demonstrated that a single, stable pre-training procedure is possible with Inductive Moment Matching (IMM). IMM offers a promising single-stage alternative to the current two-stage ``diffusion then distillation'' paradigm in visual foundational models~\citep{luo2023latent,sauer2024fast,yin2024one,zhu2024slimflow,zhou2024adversarial,lin2025diffusion}.

Given the simplicity of moment matching and its relative low popularity in the community of visual generative modeling~\citep{li2015generative}, IMM should be viewed as one point in a broader design space rather than the only alternative to diffusion distillation. Later methods have already explored related ways to learn efficient one-step or few-step generators; Section~4.3 discusses several of these alternatives in more detail. However, it is imperative to keep the network's dependence over $s$ if one wishes to employ a DDIM-style sampler.

\subsection{Multi-token prediction}

\begin{takeaway}
\textbf{Core issue.} Predicting several future tokens in parallel can reduce latency, but independently sampling those positions does not represent their joint distribution.
\end{takeaway}

Multi-token prediction (MTP) is of great interest to the language modeling community because of its potential to achieve faster inference~\citep{gloeckle2024better}, which allows efficient inference-time scaling. However, current multi-token prediction models often predict the softmax values of multiple tokens in parallel, which is a na\"{i}ve conditional independence assumption (\textit{i.e.}, na\"{i}ve Bayes). We argue that this inference design greatly limits the capacity of the model distribution and more efforts should be spent resolving this fundamental issue.

\begin{figure}[t]
\centering
\includegraphics[width=\linewidth]{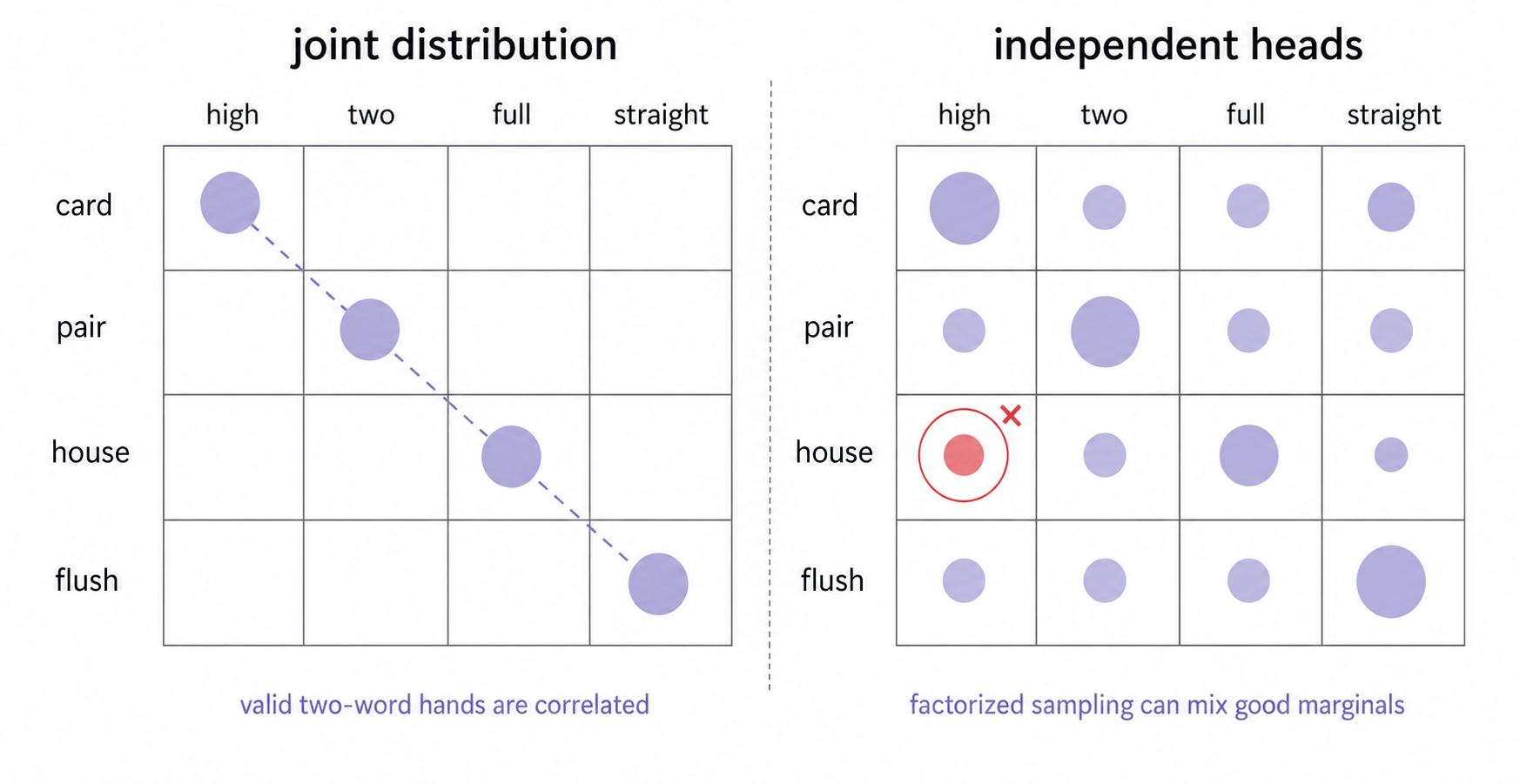}
\caption{A two-token poker-hand example. The marginal choices for each token position can all look reasonable, but independently sampling the two positions does not represent the joint distribution over valid two-word hands.}
\label{fig:mtp-poker}
\end{figure}

Consider the example of trying to predict the next two words/tokens\footnote{For simplicity of the argument, let's assume that a token is a word here.} in the sentence: ``The list of poker hands that consist of two English words are: ...''. As the list can be arbitrarily ordered, the immediate next two words can be any hand, such as ``high card'', ``two pair'', ``full house'', or ``straight flush''; there is a correlation between the two words that makes a valid hand. The multi-token prediction LLM first produces the corresponding softmax weights, and then samples the tokens independently, which may lead to unwanted combinations such as ``high house'' (Figure~\ref{fig:mtp-poker}).

\paragraph{Current workaround.}
Current MTP inference algorithms heuristically bypass this model capacity limitation. For example, Gloeckle \textit{et al.}~\citep{gloeckle2024better} use self-speculative decoding, whereas Deepseek-V3~\citep{liu2024deepseek} discards the multi-token prediction heads and uses regular next-token prediction during inference. Therefore, even though the model is misspecified in representing the joint distribution of multi-token outcomes, inference algorithms can be designed to ``correct'' this inherent capacity problem.

Notably, MTP is also relevant to design choices in diffusion-based language models~\citep{lou2023discrete,shi2024simplified,sahoo2025simple,nie2025large}. To address the capacity issue while realising latency gains, diffusion language models can first sample a few tokens in parallel, making the procedure closer to any-order autoregression~\citep{shih2022training}. They can then expand parallelism once the remaining masked tokens are closer to conditionally independent~\citep{shi2024simplified}.

\paragraph{Recent evidence.}
Recent analyses make this limitation more explicit. ParallelBench argues that parallel decoding in diffusion LLMs can ignore strong token dependencies under a factorized approximation, causing quality degradation in dependency-heavy cases~\citep{kang2025parallelbench}. Another study of masked diffusion language models finds that current models often realize less parallelism than promised and that parallel probabilistic modeling weakens inter-token dependencies~\citep{zhong2026parallelism}. Related work on truly parallel diffusion decoding argues that practical DLMs frequently collapse toward autoregressive-like decoding when trained on sequential supervision~\citep{li2026struggle}.

Other recent methods respond by predicting token dependencies for parallel unmasking~\citep{ringel2026dependency}, allowing soft self-revision to mitigate error accumulation under aggressive parallel decoding~\citep{chen2026dmax}, or distilling an autoregressive teacher through Gumbel noise to better train parallel decoders~\citep{zhang2026gumbel}. These papers support the same inference-first diagnosis: parallel decoding is only useful when the inference algorithm accounts for the dependency structure of the tokens being decoded together.

\paragraph{Open direction.}
Despite the current corrections to the inference procedure, it is interesting to ask whether we can train a well-specified multi-token prediction model without the na\"{i}ve Bayes assumption, so that we can directly sample multiple tokens during inference without rejection sampling. One challenge is to optimize through the discrete token sampling processes~\citep{jang2016categorical,grathwohl2017backpropagation}. Resolving this issue could unlock more of the state-refinement scaling potential for language models. 

\subsection{Recent evidence from flow maps and few-step distillation}

\begin{takeaway}
\textbf{Core issue.} The target-time argument above suggests that efficient few-step generation should learn the long-range transport map directly.
\end{takeaway}

\paragraph{From target time to flow maps.}
A velocity field $v_\theta(\rvx_t,t)$ defines an infinitesimal update, while a flow map directly represents a two-time transport operator from $(\rvx_t,t)$ to a target time $s$. If the model is expected to make one or few long jumps, then the two-time map is the more natural object to learn: it learns the finite transport induced by the local dynamics rather than only the local dynamics themselves~\citep{dieleman2026flowmaps}. Flow Map Matching, introduced by Boffi, Albergo, and Vanden-Eijnden, formalizes this view by learning the two-time flow map of an underlying ODE and shows that consistency models, consistency trajectory models, progressive distillation, and related neural-operator views can be interpreted as special cases~\citep{boffi2024flowmap}. This provides a theoretical language for the simple modification in Figure~\ref{fig:ddim}: conditioning on $s$ is not merely an engineering trick, but a minimal way of exposing the desired flow-map degree of freedom to the network.

\paragraph{What later work made explicit.}
The subsequent literature has moved strongly in this direction, and several later papers make the Boffi \textit{et al.} flow-map viewpoint more explicit. Their self-distillation framework derives direct training algorithms from the tangent relation between a flow map and its induced velocity field, giving a constructive recipe for building consistency models from flow maps~\citep{boffi2025selfdistillation}. Generalised Flow Maps extend the framework to Riemannian manifolds, showing how the same few-step principle can apply to geometric domains such as molecular, protein, hyperbolic, and geospatial data~\citep{davis2025generalised}. Flow-map reward guidance, stochastic flow-map alignment, and Flow Map Trajectory Tilting further suggest that the learned map can be used not only for fast integration, but also as the object through which inference-time search, guidance, and reward optimization are made efficient~\citep{holderrieth2026diamond,huang2026guide,sabour2025testtime}.

\paragraph{Shared inference shape.}
Other contemporaneous methods support the same broader lesson. Shortcut models condition on the desired step size and train a single model that can be sampled with different step budgets~\citep{frans2024shortcut}. Align Your Flow scales continuous-time flow-map distillation and reports strong few-step image and text-to-image generation, while emphasizing that flow maps avoid the degradation with increased step counts that can appear in one-step consistency models~\citep{sabour2025align}. MeanFlow similarly replaces instantaneous velocity prediction with an average-velocity object and achieves strong one-step results without a pretrained teacher~\citep{geng2025meanflows}. Terminal Velocity Matching generalizes flow matching by modeling transitions between arbitrary diffusion timesteps and regularizing behavior at the terminal time~\citep{zhou2025terminal}. Trajectory Distribution Matching (TDM) studies few-step distillation by matching both the endpoint distribution and the intermediate trajectory distribution, further reinforcing that the path taken by a few-step sampler is part of the object being learned~\citep{luo2025tdm}.

These methods are trained very differently, but they share the same underlying inference procedure: learn a long-range update rather than repeatedly applying only a local vector field. Flow maps trade local simplicity for a harder long-jump learning problem~\citep{dieleman2026flowmaps}. Still, these methods collectively validate the prediction that efficient generative pre-training should parameterize long-range inference moves directly.

\paragraph{Language modeling evidence.}
There is also early evidence that this flow-map perspective is not restricted to continuous images, but the language-modeling evidence should be separated by state space. Categorical Flow Maps define flow maps over the simplex and report strong few-step results for images, molecular graphs, and text, while Discrete Flow Maps adapt the flow-map construction directly to the geometry of the probability simplex~\citep{roos2026categorical,potaptchik2026discrete}. These works are related to continuous-space language modeling, but they are not the same object.

In the continuous direction, Continuous Diffusion for Categorical Data is an early example of diffusion for categorical data that remains continuous in both time and input space, and Flow Map Language Models distill a continuous flow language model into a flow map for few-step generation~\citep{dieleman2022continuous,lee2026flowmaplanguage}. More recently, LangFlow argues that continuous diffusion language models can become competitive with discrete diffusion by connecting embedding-space diffusion to flow matching and improving likelihood estimation, scheduling, and self-conditioning~\citep{chen2026langflow}.

By contrast, LLaDA and Block Diffusion are important masked, discrete, or hybrid diffusion language models: LLaDA uses a forward masking process and predicts masked tokens under a pre-training and instruction-tuning recipe, while Block Diffusion generates blocks autoregressively and denoises within each block~\citep{nie2025large,arriola2025block}. Together, these works suggest that the continuous-domain flow-map view and the language-modeling question are converging, but they should not be conflated: the key issue is whether the inference parameterization can represent stable, efficient long-range moves in its chosen state space.

\section{Conclusions}

The central lesson is to analyze inference before training. A DDIM-style sampler that asks for a target time without conditioning on it omits a necessary variable; a multi-token sampler that factorizes a joint distribution inherits a capacity limit. These failures are properties of the inference algorithm, not merely artifacts of the objective.

The same lens clarifies the recent literature. Flow maps, shortcut models, mean flows, Terminal Velocity Matching, Trajectory Distribution Matching, and related few-step methods differ substantially in objective design and supervision, but they share a common inference shape: learn long-range updates, not just local vector fields.

For language models, masked diffusion, block diffusion, continuous diffusion over categorical data, and flow-map language models should be compared with care because they operate in different state spaces. What matters for this paper's argument is not the label attached to the model, but whether its inference parameterization can represent the distributional move it is asked to make.

We therefore view sequence expansion and state refinement as complementary inference-time scaling axes. Autoregressive models have made sequence expansion reliable because each additional token extends a valid prefix by another normalized conditional draw. Diffusion-style and flow-map methods offer a different opportunity: they can revise an existing state and make larger parallel moves, but only when the state space, conditioning variables, and update map are specified correctly. Future generative pre-training algorithms should be judged by how efficiently they scale along these axes, and whether their inference procedures make the desired long-range moves tractable.

\section*{Acknowledgements}
We thank Jiaxin Shi, Qinsheng Zhang, Chongxuan Li, Nicholas M. Boffi, Michael S. Albergo, Eric Vanden-Eijnden, as well as members of the Luma AI research team for helpful discussions.

\bibliographystyle{alpha}
\bibliography{bib}

@string{iclr={ICLR}}

@string{siggraph={SIGGRAPH}}

@incollection{goodfellow2014generative,
	title        = {Generative Adversarial Nets},
	author       = {Goodfellow, Ian and Pouget-Abadie, Jean and Mirza, Mehdi and Xu, Bing and Warde-Farley, David and Ozair, Sherjil and Courville, Aaron and Bengio, Yoshua},
	year         = 2014,
	booktitle    = {Advances in Neural Information Processing Systems 27},
	publisher    = {Curran Associates, Inc.},
	pages        = {2672--2680},
	issn         = {1049-5258},
	editor       = {Ghahramani, Z and Welling, M and Cortes, C and Lawrence, N D and Weinberger, K Q}
}

@article{grathwohl2017backpropagation,
	title        = {Backpropagation through the Void: Optimizing control variates for black-box gradient estimation},
	author       = {Grathwohl, Will and Choi, Dami and Wu, Yuhuai and Roeder, Geoff and Duvenaud, David},
	year         = 2017,
	month        = {October},
	journal      = {arXiv preprint arXiv:1711.00123},
	archiveprefix = {arXiv},
	eprint       = {1711.00123},
	primaryclass = {cs.LG},
	arxivid      = {1711.00123}
}

@article{kingma2013auto,
	title        = {{Auto-Encoding} Variational Bayes},
	author       = {Kingma, Diederik P and Welling, Max},
	year         = 2013,
	month        = {December},
	journal      = {arXiv preprint arXiv:1312.6114v10},
	archiveprefix = {arXiv},
	eprint       = {1312.6114v10},
	primaryclass = {stat.ML},
	arxivid      = {1312.6114v10}
}

@article{rezende2015variational,
	title        = {Variational Inference with Normalizing Flows},
	author       = {Rezende, Danilo Jimenez and Mohamed, Shakir},
	year         = 2015,
	month        = {May},
	journal      = {arXiv preprint arXiv:1505.05770},
	archiveprefix = {arXiv},
	eprint       = {1505.05770},
	primaryclass = {stat.ML},
	arxivid      = {1505.05770}
}

@article{jang2016categorical,
	title        = {Categorical reparameterization with gumbel-softmax},
	author       = {Jang, Eric and Gu, Shixiang and Poole, Ben},
	year         = 2016,
	month        = {November},
	journal      = {arXiv preprint arXiv:1611.01144},
	archiveprefix = {arXiv},
	eprint       = {1611.01144},
	primaryclass = {stat.ML},
	arxivid      = {1611.01144}
}

@article{du2019implicit,
	title        = {Implicit Generation and Generalization in {Energy-Based} Models},
	author       = {Du, Yilun and Mordatch, Igor},
	year         = 2019,
	month        = {March},
	journal      = {arXiv preprint arXiv:1903.08689},
	archiveprefix = {arXiv},
	eprint       = {1903.08689},
	primaryclass = {cs.LG},
	arxivid      = {1903.08689}
}

@article{oord2016pixel,
	title        = {Pixel Recurrent Neural Networks},
	author       = {van den Oord, Aaron and Kalchbrenner, Nal and Kavukcuoglu, Koray},
	year         = 2016,
	month        = {January},
	journal      = {arXiv preprint arXiv:1601.06759},
	archiveprefix = {arXiv},
	eprint       = {1601.06759},
	primaryclass = {cs.CV},
	arxivid      = {1601.06759}
}

@article{ho2020denoising,
	title        = {Denoising Diffusion Probabilistic Models},
	author       = {Ho, Jonathan and Jain, Ajay and Abbeel, Pieter},
	year         = 2020,
	month        = {June},
	journal      = {arXiv preprint arXiv:2006.11239},
	archiveprefix = {arXiv},
	eprint       = {2006.11239},
	primaryclass = {cs.LG},
	arxivid      = {2006.11239}
}

@article{brown2020language,
	title        = {Language Models are {Few-Shot} Learners},
	author       = {Brown, Tom B and Mann, Benjamin and Ryder, Nick and Subbiah, Melanie and Kaplan, Jared and Dhariwal, Prafulla and Neelakantan, Arvind and Shyam, Pranav and Sastry, Girish and Askell, Amanda and Agarwal, Sandhini and Herbert-Voss, Ariel and Krueger, Gretchen and Henighan, Tom and Child, Rewon and Ramesh, Aditya and Ziegler, Daniel M and Wu, Jeffrey and Winter, Clemens and Hesse, Christopher and Chen, Mark and Sigler, Eric and Litwin, Mateusz and Gray, Scott and Chess, Benjamin and Clark, Jack and Berner, Christopher and McCandlish, Sam and Radford, Alec and Sutskever, Ilya and Amodei, Dario},
	year         = 2020,
	month        = {May},
	journal      = {arXiv preprint arXiv:2005.14165},
	archiveprefix = {arXiv},
	eprint       = {2005.14165},
	primaryclass = {cs.CL},
	arxivid      = {2005.14165}
}

@article{song2020denoising,
	title        = {Denoising Diffusion Implicit Models},
	author       = {Song, Jiaming and Meng, Chenlin and Ermon, Stefano},
	year         = 2020,
	month        = {April},
	journal      = {arXiv preprint arXiv:2010.02502},
	booktitle    = {International Conference on Learning Representations},
	url          = {https://arxiv.org/abs/2010.02502},
	abbr         = {ICLR 2021},
	code         = {https://github.com/ermongroup/ddim},
	count        = 37
}

@article{song2020score,
	title        = {Score-Based Generative Modeling through Stochastic Differential Equations},
	author       = {Song, Yang and Sohl-Dickstein, Jascha and Kingma, Diederik P and Kumar, Abhishek and Ermon, Stefano and Poole, Ben},
	year         = 2020,
	journal      = {arXiv preprint arXiv:2011.13456}
}

@inproceedings{li2015generative,
  title={Generative moment matching networks},
  author={Li, Yujia and Swersky, Kevin and Zemel, Rich},
  booktitle={International Conference on Machine Learning},
  pages={1718--1727},
  year={2015},
  organization={PMLR}
}

@string{ iclr        = {International Conference on Learning Representations} }

@string{ mit         = {Massachusetts Institute of Technology} }

@string{Computer = "{IEEE} Computer" }

@string{Springer = "Springer-Verlag" }

@inproceedings{chang2022maskgit,
  title={Maskgit: Masked generative image transformer},
  author={Chang, Huiwen and Zhang, Han and Jiang, Lu and Liu, Ce and Freeman, William T},
  booktitle={Proceedings of the IEEE/CVF conference on computer vision and pattern recognition},
  pages={11315--11325},
  year={2022}
}

@article{tian2025visual,
  title={Visual autoregressive modeling: Scalable image generation via next-scale prediction},
  author={Tian, Keyu and Jiang, Yi and Yuan, Zehuan and Peng, Bingyue and Wang, Liwei},
  journal={Advances in neural information processing systems},
  volume={37},
  pages={84839--84865},
  year={2025}
}

@article{li2025autoregressive,
  title={Autoregressive image generation without vector quantization},
  author={Li, Tianhong and Tian, Yonglong and Li, He and Deng, Mingyang and He, Kaiming},
  journal={Advances in Neural Information Processing Systems},
  volume={37},
  pages={56424--56445},
  year={2025}
}

@article{meng2021improved,
  title={Improved autoregressive modeling with distribution smoothing},
  author={Meng, Chenlin and Song, Jiaming and Song, Yang and Zhao, Shengjia and Ermon, Stefano},
  journal={arXiv preprint arXiv:2103.15089},
  year={2021}
}

@article{wu2023ar,
  title={Ar-diffusion: Auto-regressive diffusion model for text generation},
  author={Wu, Tong and Fan, Zhihao and Liu, Xiao and Zheng, Hai-Tao and Gong, Yeyun and Jiao, Jian and Li, Juntao and Guo, Jian and Duan, Nan and Chen, Weizhu and others},
  journal={Advances in Neural Information Processing Systems},
  volume={36},
  pages={39957--39974},
  year={2023}
}

@article{ruhe2024rolling,
  title={Rolling diffusion models},
  author={Ruhe, David and Heek, Jonathan and Salimans, Tim and Hoogeboom, Emiel},
  journal={arXiv preprint arXiv:2402.09470},
  year={2024}
}

@article{song2023consistency,
  title={Consistency models},
  author={Song, Yang and Dhariwal, Prafulla and Chen, Mark and Sutskever, Ilya},
  journal={arXiv preprint arXiv:2303.01469},
  year={2023}
}

@inproceedings{song2021accelerating,
  title={Accelerating feedforward computation via parallel nonlinear equation solving},
  author={Song, Yang and Meng, Chenlin and Liao, Renjie and Ermon, Stefano},
  booktitle={International Conference on Machine Learning},
  pages={9791--9800},
  year={2021},
  organization={PMLR}
}

@article{zhai2024normalizing,
  title={Normalizing flows are capable generative models},
  author={Zhai, Shuangfei and Zhang, Ruixiang and Nakkiran, Preetum and Berthelot, David and Gu, Jiatao and Zheng, Huangjie and Chen, Tianrong and Bautista, Miguel Angel and Jaitly, Navdeep and Susskind, Josh},
  journal={arXiv preprint arXiv:2412.06329},
  year={2024}
}

@article{kolesnikov2024jet,
  title={Jet: A Modern Transformer-Based Normalizing Flow},
  author={Kolesnikov, Alexander and Pinto, Andr{\'e} Susano and Tschannen, Michael},
  journal={arXiv preprint arXiv:2412.15129},
  year={2024}
}

@article{huang2025gan,
  title={The GAN is dead; long live the GAN! A Modern GAN Baseline},
  author={Huang, Nick and Gokaslan, Aaron and Kuleshov, Volodymyr and Tompkin, James},
  journal={Advances in Neural Information Processing Systems},
  volume={37},
  pages={44177--44215},
  year={2025}
}

@inproceedings{kang2023scaling,
  title={Scaling up gans for text-to-image synthesis},
  author={Kang, Minguk and Zhu, Jun-Yan and Zhang, Richard and Park, Jaesik and Shechtman, Eli and Paris, Sylvain and Park, Taesung},
  booktitle={Proceedings of the IEEE/CVF conference on computer vision and pattern recognition},
  pages={10124--10134},
  year={2023}
}

@article{zhu2023exploring,
  title={Exploring sparse MoE in GANs for text-conditioned image synthesis},
  author={Zhu, Jiapeng and Yang, Ceyuan and Zheng, Kecheng and Xu, Yinghao and Shi, Zifan and Shen, Yujun},
  journal={arXiv preprint arXiv:2309.03904},
  year={2023}
}

@article{liu2022flow,
  title={Flow straight and fast: Learning to generate and transfer data with rectified flow},
  author={Liu, Xingchao and Gong, Chengyue and Liu, Qiang},
  journal={arXiv preprint arXiv:2209.03003},
  year={2022}
}

@article{xu2023restart,
  title={Restart sampling for improving generative processes},
  author={Xu, Yilun and Deng, Mingyang and Cheng, Xiang and Tian, Yonglong and Liu, Ziming and Jaakkola, Tommi},
  journal={Advances in Neural Information Processing Systems},
  volume={36},
  pages={76806--76838},
  year={2023}
}

@article{liu2024deepseek,
  title={Deepseek-v3 technical report},
  author={Liu, Aixin and Feng, Bei and Xue, Bing and Wang, Bingxuan and Wu, Bochao and Lu, Chengda and Zhao, Chenggang and Deng, Chengqi and Zhang, Chenyu and Ruan, Chong and others},
  journal={arXiv preprint arXiv:2412.19437},
  year={2024}
}

@article{gloeckle2024better,
  title={Better \& faster large language models via multi-token prediction},
  author={Gloeckle, Fabian and Idrissi, Badr Youbi and Rozi{\`e}re, Baptiste and Lopez-Paz, David and Synnaeve, Gabriel},
  journal={arXiv preprint arXiv:2404.19737},
  year={2024}
}

@article{kim2023consistency,
  title={Consistency trajectory models: Learning probability flow ode trajectory of diffusion},
  author={Kim, Dongjun and Lai, Chieh-Hsin and Liao, Wei-Hsiang and Murata, Naoki and Takida, Yuhta and Uesaka, Toshimitsu and He, Yutong and Mitsufuji, Yuki and Ermon, Stefano},
  journal={arXiv preprint arXiv:2310.02279},
  year={2023}
}

@article{stern2018blockwise,
  title={Blockwise parallel decoding for deep autoregressive models},
  author={Stern, Mitchell and Shazeer, Noam and Uszkoreit, Jakob},
  journal={Advances in Neural Information Processing Systems},
  volume={31},
  year={2018}
}

@article{zhou2025inductive,
  title={Inductive Moment Matching},
  author={Zhou, Linqi and Ermon, Stefano and Song, Jiaming},
  journal={arXiv preprint},
  year={2025}
}

@inproceedings{sauer2024fast,
  title={Fast high-resolution image synthesis with latent adversarial diffusion distillation},
  author={Sauer, Axel and Boesel, Frederic and Dockhorn, Tim and Blattmann, Andreas and Esser, Patrick and Rombach, Robin},
  booktitle={SIGGRAPH Asia 2024 Conference Papers},
  pages={1--11},
  year={2024}
}

@inproceedings{yin2024one,
  title={One-step diffusion with distribution matching distillation},
  author={Yin, Tianwei and Gharbi, Micha{\"e}l and Zhang, Richard and Shechtman, Eli and Durand, Fredo and Freeman, William T and Park, Taesung},
  booktitle={Proceedings of the IEEE/CVF conference on computer vision and pattern recognition},
  pages={6613--6623},
  year={2024}
}

@inproceedings{zhu2024slimflow,
  title={Slimflow: Training smaller one-step diffusion models with rectified flow},
  author={Zhu, Yuanzhi and Liu, Xingchao and Liu, Qiang},
  booktitle={European Conference on Computer Vision},
  pages={342--359},
  year={2024},
  organization={Springer}
}

@article{zhou2024adversarial,
  title={Adversarial Score identity Distillation: Rapidly Surpassing the Teacher in One Step},
  author={Zhou, Mingyuan and Zheng, Huangjie and Gu, Yi and Wang, Zhendong and Huang, Hai},
  journal={arXiv preprint arXiv:2410.14919},
  year={2024}
}

@article{lin2025diffusion,
  title={Diffusion adversarial post-training for one-step video generation},
  author={Lin, Shanchuan and Xia, Xin and Ren, Yuxi and Yang, Ceyuan and Xiao, Xuefeng and Jiang, Lu},
  journal={arXiv preprint arXiv:2501.08316},
  year={2025}
}

@article{luo2023latent,
  title={Latent consistency models: Synthesizing high-resolution images with few-step inference},
  author={Luo, Simian and Tan, Yiqin and Huang, Longbo and Li, Jian and Zhao, Hang},
  journal={arXiv preprint arXiv:2310.04378},
  year={2023}
}

@article{salimans2022progressive,
  title={Progressive distillation for fast sampling of diffusion models},
  author={Salimans, Tim and Ho, Jonathan},
  journal={arXiv preprint arXiv:2202.00512},
  year={2022}
}

@article{shih2022training,
  title={Training and inference on any-order autoregressive models the right way},
  author={Shih, Andy and Sadigh, Dorsa and Ermon, Stefano},
  journal={Advances in Neural Information Processing Systems},
  volume={35},
  pages={2762--2775},
  year={2022}
}

@article{vincent2011connection,
  title={A connection between score matching and denoising autoencoders},
  author={Vincent, Pascal},
  journal={Neural computation},
  volume={23},
  number={7},
  pages={1661--1674},
  year={2011},
  publisher={MIT Press}
}

@article{ou2024your,
  title={Your absorbing discrete diffusion secretly models the conditional distributions of clean data},
  author={Ou, Jingyang and Nie, Shen and Xue, Kaiwen and Zhu, Fengqi and Sun, Jiacheng and Li, Zhenguo and Li, Chongxuan},
  journal={arXiv preprint arXiv:2406.03736},
  year={2024}
}

@article{sahoo2025simple,
  title={Simple and effective masked diffusion language models},
  author={Sahoo, Subham and Arriola, Marianne and Schiff, Yair and Gokaslan, Aaron and Marroquin, Edgar and Chiu, Justin and Rush, Alexander and Kuleshov, Volodymyr},
  journal={Advances in Neural Information Processing Systems},
  volume={37},
  pages={130136--130184},
  year={2025}
}

@article{shi2024simplified,
  title={Simplified and generalized masked diffusion for discrete data},
  author={Shi, Jiaxin and Han, Kehang and Wang, Zhe and Doucet, Arnaud and Titsias, Michalis},
  journal={Advances in neural information processing systems},
  volume={37},
  pages={103131--103167},
  year={2024}
}

@article{lou2023discrete,
  title={Discrete diffusion modeling by estimating the ratios of the data distribution},
  author={Lou, Aaron and Meng, Chenlin and Ermon, Stefano},
  journal={arXiv preprint arXiv:2310.16834},
  year={2023}
}

@article{hornik1989multilayer,
  title={Multilayer feedforward networks are universal approximators},
  author={Hornik, Kurt and Stinchcombe, Maxwell and White, Halbert},
  journal={Neural networks},
  volume={2},
  number={5},
  pages={359--366},
  year={1989},
  publisher={Elsevier}
}

@article{wei2022chain,
  title={Chain-of-Thought Prompting Elicits Reasoning in Large Language Models},
  author={Wei, Jason and Wang, Xuezhi and Schuurmans, Dale and Bosma, Maarten and Ichter, Brian and Xia, Fei and Chi, Ed H and Le, Quoc V and Zhou, Denny},
  journal={Advances in Neural Information Processing Systems},
  volume={35},
  pages={24824--24837},
  year={2022}
}

@article{brown2024large,
  title={Large language monkeys: Scaling inference compute with repeated sampling},
  author={Brown, Bradley and Juravsky, Jordan and Ehrlich, Ryan and Clark, Ronald and Le, Quoc V and R{\'e}, Christopher and Mirhoseini, Azalia},
  journal={arXiv preprint arXiv:2407.21787},
  year={2024}
}

@article{snell2024scaling,
  title={Scaling llm test-time compute optimally can be more effective than scaling model parameters},
  author={Snell, Charlie and Lee, Jaehoon and Xu, Kelvin and Kumar, Aviral},
  journal={arXiv preprint arXiv:2408.03314},
  year={2024}
}

@article{ma2025inference,
  title={Inference-time scaling for diffusion models beyond scaling denoising steps},
  author={Ma, Nanye and Tong, Shangyuan and Jia, Haolin and Hu, Hexiang and Su, Yu-Chuan and Zhang, Mingda and Yang, Xuan and Li, Yandong and Jaakkola, Tommi and Jia, Xuhui and others},
  journal={arXiv preprint arXiv:2501.09732},
  year={2025}
}

@article{lipman2022flow,
  title={Flow matching for generative modeling},
  author={Lipman, Yaron and Chen, Ricky TQ and Ben-Hamu, Heli and Nickel, Maximilian and Le, Matt},
  journal={arXiv preprint arXiv:2210.02747},
  year={2022}
}

@article{gretton2012kernel,
  title={A kernel two-sample test},
  author={Gretton, Arthur and Borgwardt, Karsten M and Rasch, Malte J and Sch{\"o}lkopf, Bernhard and Smola, Alexander},
  journal={The Journal of Machine Learning Research},
  volume={13},
  number={1},
  pages={723--773},
  year={2012},
  publisher={JMLR. org}
}

@article{gao2025diffusionmeetsflow,
  author = {Gao, Ruiqi and Hoogeboom, Emiel and Heek, Jonathan and Bortoli, Valentin De and Murphy, Kevin P. and Salimans, Tim},
  title = {Diffusion Meets Flow Matching: Two Sides of the Same Coin},
  year = {2024},
  journal = {The Internet},
  url  = {https://diffusionflow.github.io/}
}

@article{gat2024discrete,
  title={Discrete flow matching},
  author={Gat, Itai and Remez, Tal and Shaul, Neta and Kreuk, Felix and Chen, Ricky TQ and Synnaeve, Gabriel and Adi, Yossi and Lipman, Yaron},
  journal={Advances in Neural Information Processing Systems},
  volume={37},
  pages={133345--133385},
  year={2024}
}

@article{nie2025large,
  title={Large Language Diffusion Models},
  author={Nie, Shen and Zhu, Fengqi and You, Zebin and Zhang, Xiaolu and Ou, Jingyang and Hu, Jun and Zhou, Jun and Lin, Yankai and Wen, Ji-Rong and Li, Chongxuan},
  journal={arXiv preprint arXiv:2502.09992},
  year={2025}
}

@article{boffi2024flowmap,
  title={Flow Map Matching},
  author={Boffi, Nicholas M and Albergo, Michael S and Vanden-Eijnden, Eric},
  journal={arXiv preprint arXiv:2406.07507},
  year={2024}
}

@article{frans2024shortcut,
  title={One Step Diffusion via Shortcut Models},
  author={Frans, Kevin and Hafner, Danijar and Levine, Sergey and Abbeel, Pieter},
  journal={arXiv preprint arXiv:2410.12557},
  year={2024}
}

@article{boffi2025selfdistillation,
  title={How to Build a Consistency Model: Learning Flow Maps via Self-Distillation},
  author={Boffi, Nicholas M and Albergo, Michael S and Vanden-Eijnden, Eric},
  journal={arXiv preprint arXiv:2505.18825},
  year={2025}
}

@article{davis2025generalised,
  title={Generalised Flow Maps for Few-Step Generative Modelling on Riemannian Manifolds},
  author={Davis, Oscar and Albergo, Michael S and Boffi, Nicholas M and Bronstein, Michael M and Bose, Avishek Joey},
  journal={arXiv preprint arXiv:2510.21608},
  year={2025}
}

@article{holderrieth2026diamond,
  title={Diamond Maps: Efficient Reward Alignment via Stochastic Flow Maps},
  author={Holderrieth, Peter and Chen, Douglas and Eyring, Luca and Shah, Ishin and Anantharaman, Giri and He, Yutong and Akata, Zeynep and Jaakkola, Tommi and Boffi, Nicholas Matthew and Simchowitz, Max},
  journal={arXiv preprint arXiv:2602.05993},
  year={2026}
}

@article{huang2026guide,
  title={How to Guide Your Flow: Few-Step Alignment via Flow Map Reward Guidance},
  author={Huang, Jerry Y and Lin, Justin and Shah, Sheel and Nair, Kartik and Boffi, Nicholas M},
  journal={arXiv preprint arXiv:2604.27147},
  year={2026}
}

@article{sabour2025align,
  title={Align Your Flow: Scaling Continuous-Time Flow Map Distillation},
  author={Sabour, Amirmojtaba and Fidler, Sanja and Kreis, Karsten},
  journal={arXiv preprint arXiv:2506.14603},
  year={2025}
}

@article{geng2025meanflows,
  title={Mean Flows for One-Step Generative Modeling},
  author={Geng, Zhengyang and Deng, Mingyang and Bai, Xingjian and Kolter, J Zico and He, Kaiming},
  journal={arXiv preprint arXiv:2505.13447},
  year={2025}
}

@article{zhou2025terminal,
  title={Terminal Velocity Matching},
  author={Zhou, Linqi and Parger, Mathias and Haque, Ayaan and Song, Jiaming},
  journal={arXiv preprint arXiv:2511.19797},
  year={2025}
}

@article{luo2025tdm,
  title={Learning Few-Step Diffusion Models by Trajectory Distribution Matching},
  author={Luo, Weijian and Zheng, Tianyang and Zhang, Chen and Chen, Yibing and Wu, Zhengyang and Lin, Yuanhao and Wang, Zhendong and Li, Yongming and Zhang, Hao and Lu, Jianfeng},
  journal={arXiv preprint arXiv:2503.06674},
  year={2025}
}

@article{roos2026categorical,
  title={Categorical Flow Maps},
  author={Roos, Daan and Davis, Oscar and Eijkelboom, Floor and Bronstein, Michael and Welling, Max and Ceylan, {\.I}smail {\.I}lkan and Ambrogioni, Luca and van de Meent, Jan-Willem},
  journal={arXiv preprint arXiv:2602.12233},
  year={2026}
}

@article{arriola2025block,
  title={Block Diffusion: Interpolating Between Autoregressive and Diffusion Language Models},
  author={Arriola, Marianne and Gokaslan, Aaron and Chiu, Justin T and Yang, Zhihan and Qi, Zhixuan and Han, Jiaqi and Sahoo, Subham Sekhar and Kuleshov, Volodymyr},
  journal={arXiv preprint arXiv:2503.09573},
  year={2025}
}

@article{chen2026langflow,
  title={LangFlow: Continuous Diffusion Rivals Discrete in Language Modeling},
  author={Chen, Yuxin and Liang, Chumeng and Sui, Hangke and Guo, Ruihan and Cheng, Chaoran and You, Jiaxuan and Liu, Ge},
  journal={arXiv preprint arXiv:2604.11748},
  year={2026}
}

@article{gao2025dar,
  title={D-AR: Diffusion via Autoregressive Models},
  author={Gao, Ziteng and Shou, Mike Zheng},
  journal={arXiv preprint arXiv:2505.23660},
  year={2025}
}

@article{chen2024diffusionforcing,
  title={Diffusion Forcing: Next-Token Prediction Meets Full-Sequence Diffusion},
  author={Chen, Boyuan and Monso, Diego and Du, Yilun and Simchowitz, Max and Tedrake, Russ and Sitzmann, Vincent},
  journal={arXiv preprint arXiv:2407.01392},
  year={2024}
}

@article{kang2025parallelbench,
  title={ParallelBench: Understanding the Trade-offs of Parallel Decoding in Diffusion LLMs},
  author={Kang, Wonjun and Galim, Kevin and Oh, Seunghyuk and Lee, Minjae and Zeng, Yuchen and Zhang, Shuibai and Hooper, Coleman and Hu, Yuezhou and Koo, Hyung Il and Cho, Nam Ik and Lee, Kangwook},
  journal={arXiv preprint arXiv:2510.04767},
  year={2025}
}

@article{zhong2026parallelism,
  title={Parallelism and Generation Order in Masked Diffusion Language Models: Limits Today, Potential Tomorrow},
  author={Zhong, Yangyang and Gu, Yanmei and Zang, Zhengqing and Li, Xiaomeng and Ding, Yuqi and Jia, Xibei and Shen, Yuting and Lan, Zhenzhong and Zhu, Liwang and Liu, Weiping and Zhou, Junlin and Liu, Haisheng and Yu, Zhong Xin and Luo, Pengxin and Qi, Donglian and Yan, Yunfeng and Zhao, Junbo},
  journal={arXiv preprint arXiv:2601.15593},
  year={2026}
}

@article{li2026struggle,
  title={Why Diffusion Language Models Struggle with Truly Parallel (Non-Autoregressive) Decoding?},
  author={Li, Pengxiang and Muhtar, Dilxat and Yin, Lu and Chen, Tianlong and Liu, Shiwei},
  journal={arXiv preprint arXiv:2602.23225},
  year={2026}
}

@article{ringel2026dependency,
  title={Dependency-Guided Parallel Decoding in Discrete Diffusion Language Models},
  author={Ringel, Liran and Ali, Ameen and Romano, Yaniv},
  journal={arXiv preprint arXiv:2604.02560},
  year={2026}
}

@article{chen2026dmax,
  title={DMax: Aggressive Parallel Decoding for dLLMs},
  author={Chen, Zigeng and Fang, Gongfan and Ma, Xinyin and Yu, Ruonan and Wang, Xinchao},
  journal={arXiv preprint arXiv:2604.08302},
  year={2026}
}

@article{zhang2026gumbel,
  title={Gumbel Distillation for Parallel Text Generation},
  author={Zhang, Chi and Hu, Xixi and Liu, Bo and Liu, Qiang},
  journal={arXiv preprint arXiv:2603.22216},
  year={2026}
}

@misc{dieleman2026flowmaps,
  author={Dieleman, Sander},
  title={Learning the Integral of a Diffusion Model},
  url={https://sander.ai/2026/05/06/flow-maps.html},
  year={2026}
}

@article{sabour2025testtime,
  title={Test-time Scaling of Diffusions with Flow Maps},
  author={Sabour, Amirmojtaba and Albergo, Michael S. and Domingo-Enrich, Carles and Boffi, Nicholas M. and Fidler, Sanja and Kreis, Karsten and Vanden-Eijnden, Eric},
  journal={arXiv preprint arXiv:2511.22688},
  year={2025}
}

@article{lee2026flowmaplanguage,
  title={Flow Map Language Models: One-step Language Modeling via Continuous Denoising},
  author={Lee, Chanhyuk and Yoo, Jaehoon and Agarwal, Manan and Shah, Sheel and Huang, Jerry and Raghunathan, Aditi and Hong, Seunghoon and Boffi, Nicholas M. and Kim, Jinwoo},
  journal={arXiv preprint arXiv:2602.16813},
  year={2026}
}

@article{dieleman2022continuous,
  title={Continuous Diffusion for Categorical Data},
  author={Dieleman, Sander and Sartran, Laurent and Roshannai, Arman and Savinov, Nikolay and Ganin, Yaroslav and Richemond, Pierre H. and Doucet, Arnaud and Strudel, Robin and Dyer, Chris and Durkan, Conor and Hawthorne, Curtis and Leblond, R{\'e}mi and Grathwohl, Will and Adler, Jonas},
  journal={arXiv preprint arXiv:2211.15089},
  year={2022}
}

@article{potaptchik2026discrete,
  title={Discrete Flow Maps},
  author={Potaptchik, Peter and Yim, Jason and Saravanan, Adhi and Holderrieth, Peter and Vanden-Eijnden, Eric and Albergo, Michael S.},
  journal={arXiv preprint arXiv:2604.09784},
  year={2026}
}
\end{document}